%% file: main.tex
\documentclass[10pt,conference]{IEEEtran}
\usepackage{textcomp}
\usepackage[super]{nth}
\usepackage{cite}

\ifCLASSINFOpdf
   \usepackage[pdftex]{graphicx}
   \graphicspath{images}
\else
   \usepackage[dvips]{graphicx}
   \graphicspath{images}
\fi
\usepackage{algorithm2e}
\SetAlFnt{\small}
\ifCLASSOPTIONcompsoc
 \usepackage[caption=false,font=normalsize,labelfont=sf,textfont=sf]{subfig}
\else
 \usepackage[caption=false,font=footnotesize]{subfig}
\fi

\usepackage{stfloats}
\usepackage{url}

\input{acrons.tex}

\begin{document}
%
\title{Deep Reinforcement Learning in Lane Merge Coordination for Connected Vehicles}
%
%
%

\author{\IEEEauthorblockN{Omar Nassef}
\IEEEauthorblockA{\textit{Department of Informatics} \\
\textit{King's College London}\\
London, England \\
omar.nassef@kcl.ac.uk}
\and
\IEEEauthorblockN{Luis Sequeira}
\IEEEauthorblockA{\textit{Department of Informatics} \\
\textit{King's College London}\\
London, England \\
luis.sequeira@kcl.ac.uk}
\and
\IEEEauthorblockN{Elias Salam}
\IEEEauthorblockA{\textit{Department of Software Engineering} \\
\textit{Orange Labs Services}\\
Paris, France \\
elias.salam@orange.com}
\and
\IEEEauthorblockN{Toktam Mahmoodi}
\IEEEauthorblockA{\textit{Department of Informatics} \\
\textit{King's College London}\\
London, England \\
toktam.mahmoodi@kcl.ac.uk}
}

%
%

\markboth{Journal of \LaTeX\ Class Files,~Vol.~14, No.~8, August~2015}%
{Shell \MakeLowercase{\textit{et al.}}: Bare Demo of IEEEtran.cls for IEEE Journals}
%




\maketitle

\begin{abstract}
In this paper, a framework for lane merge co-ordination is presented utilising a centralised system, for connected vehicles. The delivery of trajectory recommendations to the connected vehicles on the road is based on a \textit{Traffic Orchestrator} and a \textit{Data Fusion} as the main components. Deep Reinforcement Learning and data analysis is used to predict trajectory recommendations for connected vehicles, taking into account unconnected vehicles for those suggestions. The results highlight the adaptability of the \textit{Traffic Orchestrator}, when employing \textit{Dueling Deep Q-Network} in an unseen real world merging scenario. A performance comparison of different reinforcement learning models and evaluation against \gls{kpi} are also presented.
\end{abstract}


%
\IEEEpeerreviewmaketitle

\section{Introduction}
\label{Introduction}
\input{intro.tex}

\section{State of the Art}
\label{State of the Art}
\input{soa.tex}

\section{Architecture and System Model} 
\label{Architecture and System Model}
\input{model2}
\input{to.tex}

\input{data}

\section{Results}
\label{Results}
\input{results.tex}

\section{Conclusion}
\input{conclusion.tex}

\label{Conclusion}


%



\section*{Acknowledgment}

This work has been performed in the framework of the H2020 project 5GCAR co-funded by the EU. The views expressed are those of the authors and do not necessarily represent the project. The consortium is not liable for any use that may be made of any of the information contained therein. This work is also partially funded by the EPSRC INITIATE EP/P003974/1 and The UK Programmable Fixed and Mobile Internet Infrastructure.

\ifCLASSOPTIONcaptionsoff
  \newpage
\fi



\bibliographystyle{IEEEtran}
\bibliography{references}
\end{document}

%% file: acrons.tex
\usepackage{multicol}
\usepackage{multirow}
\usepackage{amssymb}
\usepackage{colortbl}
\usepackage{tabularx}
\usepackage{siunitx}
\usepackage{lscape}
\usepackage{pdflscape}
\usepackage{booktabs}
\usepackage{lipsum}
\usepackage{gensymb}
\usepackage{amsmath}
\usepackage[]{algorithm2e}
\usepackage{float}
\usepackage[acronym,nomain,nonumberlist]{glossaries}
\usepackage{subfig}
\definecolor{row1}{rgb}{0.95,0.95,0.95}
\definecolor{row2}{rgb}{0.85,0.85,0.85}
\newacronym{5g}{$5G$}{$5^{th}$ Generation Mobile Network} 
\newacronym{cacc}{CACC}{Cooperative Adaptive Cruise Control}
\newacronym{etsi}{ETSI}{European Telecommunications Standards Institute}
\newacronym{it}{IT}{Information Technologies}
\newacronym{its}{ITS}{Intelligent Transport System} 
\newacronym{iot}{IoT}{Internet of Things}
\newacronym{mec}{MEC}{Multi-Access Edge Computing}
\newacronym{ran}{RAN}{Radio Access Network}
\newacronym{sdn}{SDN}{Software Defined Networking}
\newacronym{v2x}{V2X Gateway}{Vehicle-to-Everything}
\newacronym{v2e}{V2X}{Vehicle-to-Everything}
\newacronym{5gaa}{$5$GAA}{$5G$ Automotive Association}
\newacronym{json}{JSON}{JavaScript Object Notation}
\newacronym{tcp}{TCP}{Transmission Control Protocol}
\newacronym{udp}{UDP}{User Datagram Protocol}
\newacronym{rud}{RUD}{Road User Description}
\newacronym{id}{ID}{Identification}
\newacronym{rmse}{RMSE}{Root-Mean-Square Error}
\newacronym{gui}{GUI}{Graphical User Interface}
\newacronym{gdm}{GDM}{Global Dynamic Map}
\newacronym{uuid}{UUID}{Universally Unique Identifier}
\newacronym{lstm}{LSTM}{Long Short-Term Memeory}
\newacronym{rl}{RL}{Reinforcement Learning}
\newacronym{rfc}{RFC}{Random Forest Classifier}
\newacronym{dqn}{DQN}{Deep Q-Network}
\newacronym{ngsim}{NGSIM}{Next Generation Simulation Model}
\newacronym{to}{TO}{Traffic Orchestrator}
\newacronym{ddqn}{Double-DQN}{Double Deep Q-Network}
\newacronym{dueling-dqn}{Dueling DQN}{Dueling Deep Q-Network}
\newacronym{sarsa}{SARSA}{State-Action-Reward-State-Action}
\newacronym{kpi}{KPI}{Key Performance Indicator}
\newacronym{gelf}{GELF}{Graylog Extended Log Format}
\newacronym{df}{DF}{Data Fusion}
\newacronym{ecdf}{ECDF}{Empirical Cumulative Distribution Function}
\newacronym{rtt}{RTT}{Round Trip Time}
\newacronym{ai}{AI}{Artificial Intelligence}

%% file: intro.tex
\gls{its} enables the generation of extensive and detailed data relating to vehicles in addition to the environment of their operation. This data can be used to generate meaningful information to provide a better transportation experience. Associations such as the \gls{etsi} and \gls{5gaa} have promoted the use of cellular \gls{v2e} communications in order to enhance road safety, traffic efficiency, reduce environmental issues and energy costs \cite{its4}. A connected vehicle is capable of transmitting and receiving information to increase the consciousness and recognition of a driving agent. The necessary data can be transmitted using Vehicle-to-Vehicle and Vehicle-to-Network communications to a central system for manoeuvre generation, traffic analysis and many other use cases. Due to the advancements in \gls{5g} and \textit{\gls{v2e}}, there are many use cases that are under research and development \cite{its2}: automated overtake, co-operative collision avoidance, high density platooning and lane merging. 

In this paper, we focus on a lane merging scenario involving a vehicle merging onto a carriageway between a following and preceding vehicle. A coordination model utilising a centralised system is presented and analysed. The platform delivers trajectory recommendations to connected vehicles through the use of \textit{\gls{rl}}, accounting for all surrounding vehicles (i.e., connected and unconnected). Time-critical variables including location, speed and acceleration are used as input variables to the deep reinforcement learning model. Furthermore, various approaches to \textit{\gls{rl}} algorithms are evaluated to ascertain whether a merging vehicle can execute a manoeuvre safely, accompanied with the optimal model tested on real connected vehicles on a test track. The contributions presented in this paper include:

\begin{itemize}
    \item A \textit{\gls{to}} model based on a centralised system, that delivers trajectory recommendations to connected vehicles.
    \item Performance evaluation of different reinforcement learning approaches, when assessed in a real lane merge scenario.
    \item A \textit{\gls{df}} model that synergies the centralised micro service oriented architecture, delivering descriptors of connected and unconnected vehicles to the \textit{\gls{to}}.
\end{itemize}

Two different algorithms written in Pytorch \cite{pytorch} are presented in this work and thoroughly explored: \gls{dqn} and \gls{dueling-dqn}. The \gls{dueling-dqn} showed the most optimal results, providing human-like trajectories with very low bias. The inter vehicle distance, acceleration, individual positions and manoeuvre distance in trajectory recommendation are evaluated extensively to deduce the performance of incorporating such model.

The remainder of this paper is organised in the following way. Section \ref{State of the Art} provides a state of the art and Section \ref{Architecture and System Model} presents the architecture of the system model. The analysis of the the deep reinforcement learning algorithms is explored in section \ref{Results} on the data-set it was trained on as well as a real world scenario with the accompanying communication results. Finally, we draw conclusions and present future works in Section \ref{Conclusion}.

%% file: soa.tex
Using machine learning is by no means a new approach to tackle lane merge prediction. An approach presented in \cite{lm9}, utilised a representation of the on-road environment (Dynamic Probabilistic Drivability Map). The automotive test bed included cameras, radars and lidar sensing delivering cost effective recommendations based on dynamic programming. The theoretical formulation of this work was tested with data from $40$ real-world merges. Although the approach is considered early stage. In \cite{1m12}, the authors consider a transition stage in the path to fully autonomous transportation, with mixed-autonomy driving. The mixed-autonomy driving is considered as a collaboration of vehicles adopting a Nash Equilibrium state, to ensure that the collective reward for lane merging is optimal. The approach simulated the role of a driver via a keyboard. In \cite{lm7}, a work-in-progress for an on-ramp merge driving policy using \gls{lstm} architecture with Deep Q-Learning was presented. The scenario considers an on-ramp merging involving three vehicles: the merging vehicle and two vehicles on the mainline. A total of $9$ variables are used: for the merging vehicle, $5$ variables describe its driving state (speed, position, heading angle, and distances to the right and left lane). For the other two vehicles, only speeds and positions are known. The algorithm has not been verified or validated. A deep \textit{\gls{rl}} approach was adopted in \cite{autonomousDrivingRl}, handling input values from camera and laser sensor the vehicle owns with an embedded GPU for decision making. The work also proposes a monolith architecture embedded in the vehicle, and does not consider a micro service approach in which the connected and unconnected vehicles can co-exist in the same scenario. Lastly, every vehicle that includes the equipment specified would have to generate the calculations multiple times, making it more costly albeit running in real-time. Research has also been carried out about the functionality and challenges of incorporating Deep Learning into vehicles. Work in \cite{deepRlForVehicles} discusses the strict assessment that needs to be undertaken before the use of Deep Learning can be considered and commercialised in autonomous vehicles. Challenges included dataset completeness, Neural Network implementation and the transfer of learning. These challenges remain to this day and are great hurdles when designing and implementing a Deep Learning approach for vehicles. It is clear that reproducing similar results in different environments from research papers seldom work. Both intrinsic (e.g. hyper-parameters) and extrinsic factors (e.g. environment) influence the performance of the agent albeit using the same approach that papers have undertaken. Therefore, suggesting that the use of \textit{\gls{rl}} is experimental and relative to the scenario that the agent operates as also seen by work presented in \cite{RLPerfomance}.

%% file: model2.tex
\subsection{Lane merge coordination}

\begin{figure}[!t]
    \centering
    \includegraphics[width=0.49\textwidth,height=6.8cm]{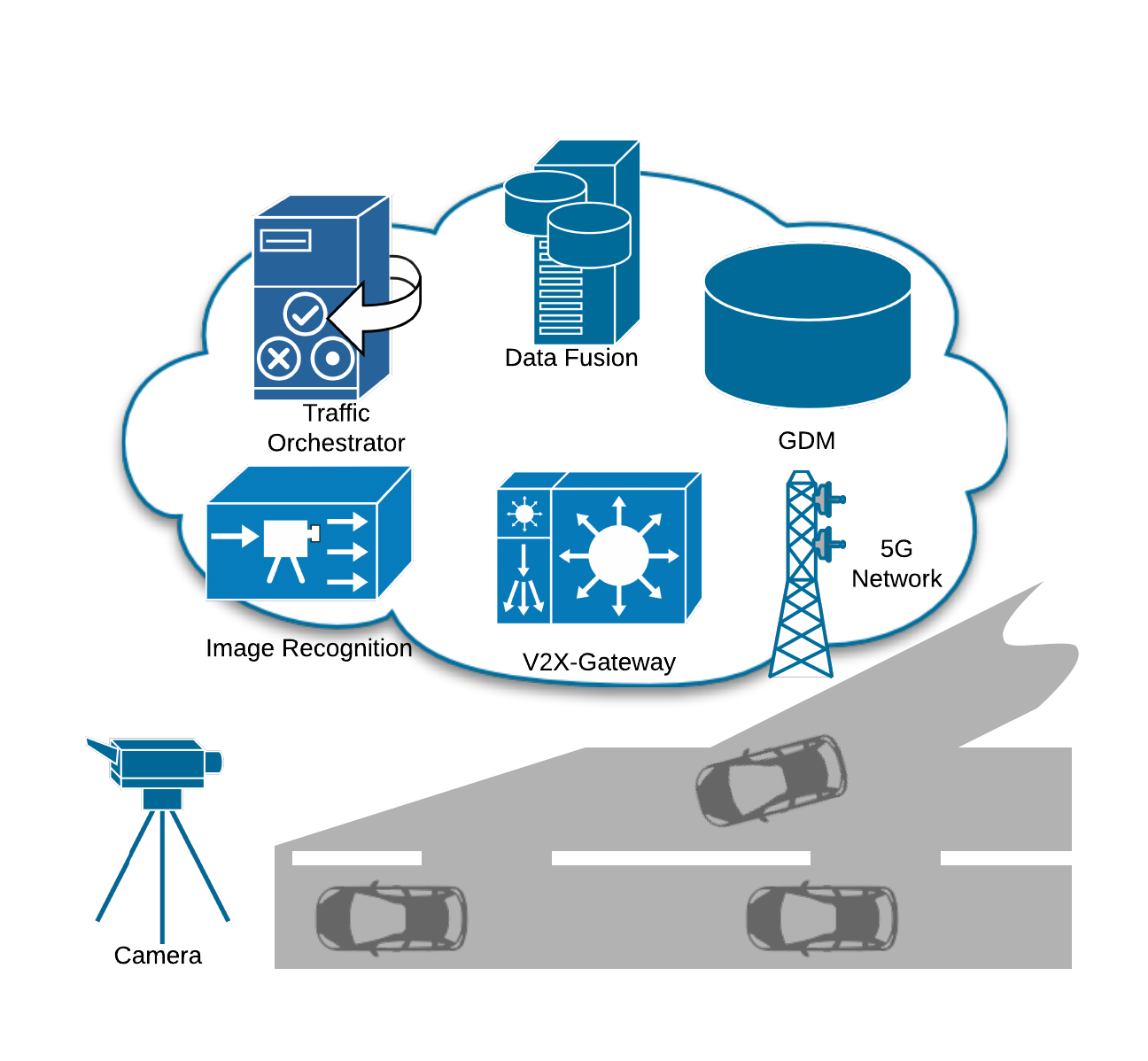}
    \caption{Lane merge coordination scenario.}
    \label{fig:LaneMergeCoordination}
\end{figure}

The lane merge scenario examined in this work is depicted in Fig. $\ref{fig:LaneMergeCoordination}$. A connected vehicle will attempt to merge onto a main lane where connected and unconnected vehicles are present. Through an edge-cloud approach, bespoke trajectory recommendations are determined and sent by central coordination mechanism to connected vehicles. This proposed architecture brings by the ability to aggregate data for various algorithms that are part of the lane merge proactively, simultaneously being able to react to sudden road changes. The proposed architecture further allows for easy scalability with respect to coordination of vehicles. Five distinct components facilitate the lane merge coordination: a \textit{\gls{v2x}}, an \textit{Image recognition} system, a \textit{\gls{gdm}}, a \textit{\gls{df}} and a \textit{\gls{to}}.

The \textit{\gls{v2x}} is responsible for forwarding messages to the various applications and interfaces in the architecture. The \textit{\gls{v2x}} acts as a communication medium that connects the interfaces and applications to connected vehicles based on a message exchanging approach. This method of communication occurs across a mobile network. Applications must subscribe to the \textit{\gls{v2x}} to receive messages about vehicular features and trajectory information. The mobile network seeks to maintain a set of baseline requirements. The up-link per vehicle should be, at least, $320kbps$. Furthermore, the down-link per vehicle should be, at least, $4.7Mbps$. The end-to-end latency requirement should not exceed $30ms$.

An \textit{Image recognition} system \cite{kai} collects information about all the vehicles on the road in a specified area. This information includes the localisation and trajectory-based parameters attributed to a specific road user, given by a \gls{rud}. Information about connected and unconnected vehicles are collected and processed sending all the information to the \gls{v2x}, which in turn forwards the messages to the \textit{\gls{gdm}}. The \textit{\gls{gdm}} stores environmental information about connected and unconnected vehicles in a database. This information is delivered from the \textit{\gls{v2x}} system. The \textit{\gls{gdm}} ensures that stored \glspl{rud} are up to date. The \textit{\gls{df}} provides a synchronisation mechanism for \glspl{rud} originating from different sources (e.g., one from the \textit{Image recognition} system and a connected vehicle in a closely localised time frame, respectively). The \textit{\gls{df}} sends the information to applications that are subscribed to a specific location boundary. Additionally, it includes the monitoring and evaluation platform to assess communication \glspl{kpi}.

The \textit{\gls{to}} will store and process environmental factors about connected and unconnected vehicles to give rise to trajectories for connected vehicles. The \textit{\gls{to}} needs to consider time-critical variables such as the timestamp of the vehicle location, the speed of the vehicle and the vehicle-specific dimensions. Once the \textit{\gls{to}} provides a coordinated trajectory recommendation for a single or set of road users, which will then be sent to the \textit{\gls{v2x}} forwarding them to the connected vehicles. The connected vehicles have the choice to either accept, reject or abort the recommendation. This feedback information is supplied by the connected vehicles to the \textit{\gls{v2x}}. The feedback can be used to recalculate trajectory recommendations. 

Unconnected vehicles are not able to communicate with the \textit{\gls{to}} and they cannot interpret or use trajectory recommendations. However the lane merge coordination is aware of unconnected vehicles by means of the \textit{Image recognition} system. This \textit{Image recognition} system provides the \textit{\gls{gdm}} with the \glspl{rud} to be stored. The road user information will be requested by the \textit{\gls{to}} to create trajectory recommendations. To this end, a set of messages need to be defined for communicating all the components within the lane merge coordination. Messages used in the communication will employ a common message formatting based on \gls{json}. This allows to communicate human-readable text, that can be received and processed in any software component.

The novelty of the the \gls{to} stems from the greater ability to generalise to different lane merge scenarios as a result of incorporating deep learning models. The \gls{to} architecture is uniquely minimal, relying on the lower levels of the architecture e.g. \textit{Image recognition} to provide status of the road and vehicles. This places a deeper focus on optimisation of computational and time resources with regards to the \gls{to} aspect. In this paper, we focus on the design, implementation and evaluation of the \textit{\gls{to}} and the \textit{\gls{df}}. The model in Fig. \ref{fig:LaneMergeCoordination} was implemented using the \textit{Image recognition} system from \cite{kai} and for the \textit{GDM}, the \textit{V2X-Gateway} and the $5G$ \textit{Network} the work from \cite{v2x_systems} was used. 

%% file: to.tex
\begin{figure*}[!t]
	\centering
	\includegraphics[width=\textwidth,height=9.5cm]{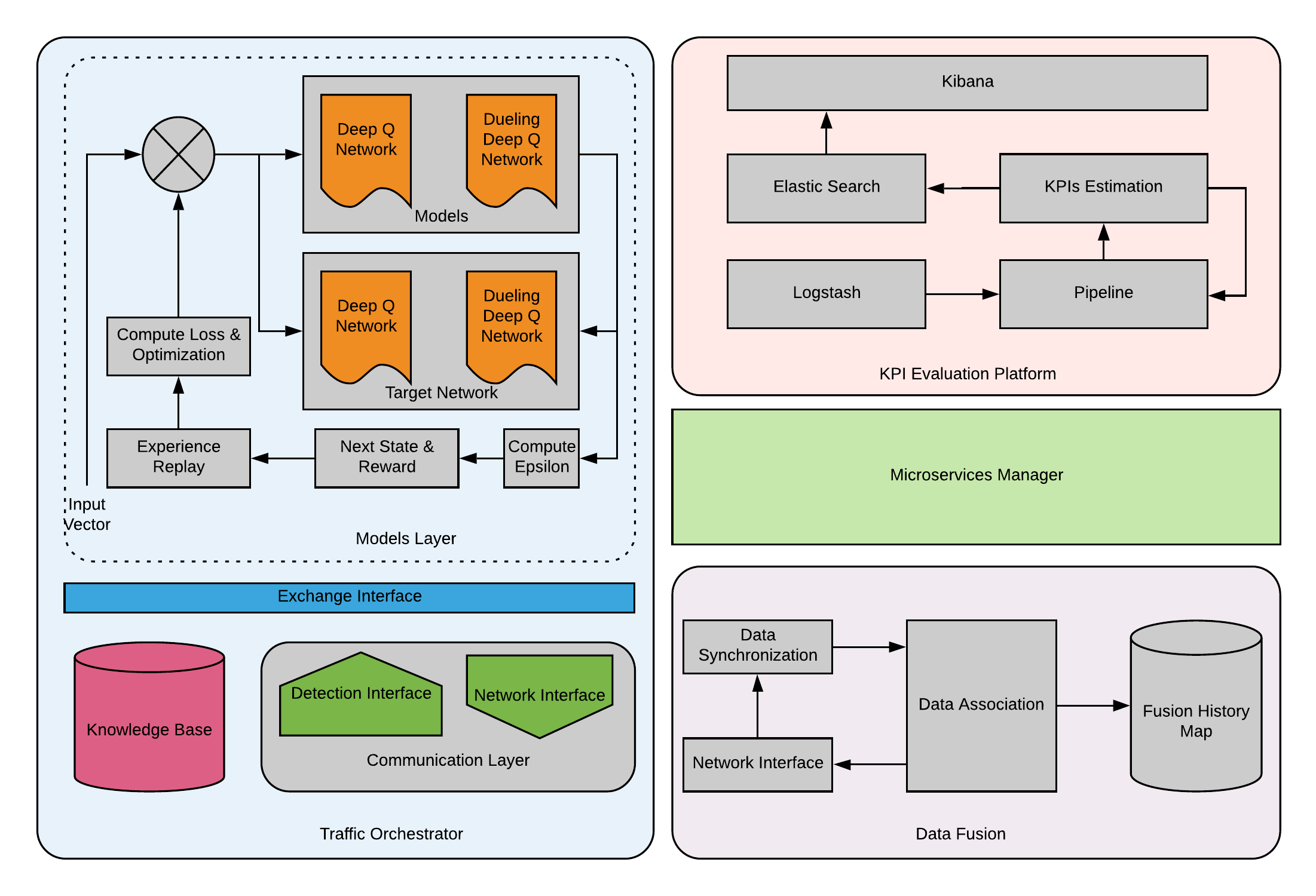}
	\caption{Proposed \textit{\gls{to}} architecture.}
	\label{proposed_architecture}
\end{figure*}

\subsection{Traffic Orchestrator}

The \textit{\gls{to}} must demonstrate a level of safety concern and overall reliability as delved in \cite{lm13}. The proposed architecture for the \textit{\gls{to}} system is presented in Fig. $\ref{proposed_architecture}$. The main purpose of the \textit{Detection Interface} is to, receiving any data being sent, over a \gls{tcp} connection, from the \textit{\gls{v2x}}. The \textit{Detection Interface} also acts as an intermediate filter that will read \gls{json} strings and process the \gls{json} messages into more compact and efficient structure to be used by the \textit{\gls{to}}. Similarly, the \textit{Network Interface} will act as a filter that will convert and translate information within the \textit{\gls{to}}, to \gls{json} messages to be fed into the \textit{\gls{v2x}}.

A \textit{Knowledge Base} has been designed to store the information sent to the \textit{\gls{to}}.Where up-to-date \glspl{rud}, is maintained to guarantee that a manoeuvre recommendation is calculated based on all current road-environment knowledge. The \textit{Knowledge Base}, mimics the access and modification functions of a typical database containing only the \glspl{rud} that the \textit{\gls{gdm}} has most recently transmitted as rows. The knowledge base is able to insert and remove \gls*{rud} to represent the most recent environmental snapshot. It also provides access to \glspl*{rud} being stored in order to query certain conditions and provide manoeuvre recommendations providing a search function allowing the retrieval of a \gls*{rud} by their \gls*{uuid}.

The \textit{Exchange Interface} has been designed to have two responsibilities: execute the \textit{\gls{to}} application and mediate the flow of information across all interfaces in the \textit{\gls{to}}. The \textit{Exchange Interface} takes structured data from the \textit{Detection Interface} and appropriately forms the data into entities. These entities can then be reused throughout the rest of the system in a consistent manner. This component directly interfaces with the \textit{Knowledge Base} and will collect structured \glspl{rud}. Another functionality of the \textit{Exchange Interface} is to provide access for consistent communication methods, allowing different learning algorithms to run on the \gls{to}. There are two major design factors with respect to a trajectory recommendation: $1)$ Safety distance from all cars on the highway; this is to ensure that the cars keep the safety breaking distance at all times, $2)$ Positioning and acceleration values of the connected vehicle in comparison with the values of all other vehicles on the road. Therefore, the motivation of the \textit{\gls{to}}, is to provide positional coordinates as well as acceleration and speed values to the connected vehicle to give a path to follow for a merge. The \textit{\gls{to}} passes instructions to other connected vehicles creating a multi-agent solution that benefits the interest of every vehicle on the road. In order for the \textit{\gls{to}} to communicate with the other components in the stack, a containerisation approach is adopted ensuring, cross platform compatibility and the ability to communicate with ease.

The \gls{rl} models utilise two different data-sets collected by the Federal Highway Administration Research and Technology - Coordinating, Developing, and Delivering Highway Transportation Innovations, on two American motorways: Interstate $80$ Freeway ($I-80$) and $US-101$. Both models require coordinates, speed, acceleration, heading and size of $3$ vehicles as input variables. The three vehicles correspond to the merging, preceding and following vehicle. The reward function have been designed to take into account the inverse distance to the merging point (position that conforms to safety measures and is in between the preceding and following vehicle), inverse of speed and acceleration. This is to ensure, a steady lane merge. The negative reward reflects the positive reward albeit with a negative sign. The actions permitted by the \gls{rl} model are: accelerate, decelerate, turn right, turn left and do nothing.

%% file: data.tex
\subsection{Data Fusion}

The \textit{\gls{df}} is responsible for updating the \textit{\gls{gdm}} data with the latest road user descriptions and avoids having a duplicated road user coming directly from connected vehicles and from the \textit{Image recognition system}. It is also responsible for enhancing the \glspl{rud} accuracy by combining most accurate values from each source, i.e the acceleration from the camera system is less precise than the one provided by the connected vehicles. The \textit{\gls{df}} consists of four different components: the \textit{Network Interface} is responsible for interacting with other components. It receives \glspl{rud} from the \textit{\gls{v2x}}, deserialises the messages and forwards them to the \textit{Data Synchronisation}. It also handles the fused and corrected descriptions from the \textit{Data Association}, serialises and sends them to the \textit{\gls{gdm}}. It receives \glspl{rud} from the \textit{\gls{v2x}}, deserialises the messages and forwards them to the \textit{Data Synchronisation}. The \textit{\gls{df}} also receives the fused and corrected descriptions from the \textit{Data Association}, serialises and sends them to the \textit{\gls{gdm}}. The \textit{Data Synchronisation} aims to synchronise received \gls{rud} in time. It collects all received descriptions during a certain period ($100 ms$) and then extrapolates each one to the same temporal reference. Given that the period is small considering the speed of the objects a uniformly accelerated rectilinear motion was found sufficient. It updates the coordinates of each object and its timestamp. The \textit{Data Association} matches objects detected by the camera system with connected vehicles. For each object detected by the camera system it first checks if the object is already matched with another object in the \textit{Fusion History Map}. Then, it raises or lowers the confidence according to the Euclidean distance and the angle between them. The \textit{Fusion History Map} stores the history of matched objects. For each matched objects it stores the following information: last seen timestamp, camera detected \gls{uuid}, connected vehicles \gls{uuid}, confidence level. The history map is cleaned every few seconds based on the last seen timestamp.

\subsection{\gls{kpi} Evaluation Platform and Micro-Services Manager}

The aim of the \gls{kpi} Evaluation Platform is firstly to ease monitoring the overall system in real time by aggregating every component logs in a single and easy to search platform and secondly evaluate software and network \glspl{kpi}, for example delays and reliability. The platform consists of three components: The collector receives logs from different components by exposing a network interface. The received messages are parsed, formatted and enriched before forwarding them to the database. The database is used to store the messages gathered by the collector and offers a query language to explore the data and compute \glspl{kpi}. The data visualisation is a GUI that allows users to monitor the database data in real time. It offers the ability to explore raw data and to create charts and dashboards.

The \textit{Elastic stack} is a set of well integrated open source components designed for this purpose with \textit{Logstash} for data collection, \textit{Elasticsearch} to store and query data and \textit{Kibana} for visualisation. The Micro-Services Manager enables connectivity among independent components in a scalable solution, providing a central logging system for all the components that can be monitored for further manipulation and analysis.

%% file: results.tex
To analyse the effectiveness of the lane merge coordination, this section focuses on two integral testing phases: a performance evaluation of the \gls{rl} models and a set of real world tests using connected vehicles. The purpose of the performance evaluation is to determine the optimal \gls{rl} model to be used for the \textit{\gls{to}} on real tests. In the real world tests, the \textit{\gls{to}} is predicting live trajectory recommendations to connected vehicles.

\subsection{Performance Evaluation Tests}

Two different \gls{rl} models (i.e, \gls{dqn} and \gls{dueling-dqn}) were trained using the data-set described in section \ref{Architecture and System Model}. This data-set was split into $3$ subsets: training, testing and validation where each of them with $70\%$, $20\%$ and $10\%$ of the size of the original data-set respectively. The training subset contains $10^5$ merging instances where each merging instance is approximately $70$ data points that represent a merging scenario. The merging scenarios are randomly selected from the data-set, but the data points are iterated over chronologically to provide a logical merging instance. The model predicts and allocates a trajectory recommendation to connected vehicles for a successful merge. The validation subset was used to tune the model hyper parameters (i.e., model layer number and size), as well as physics for the Newtonian actions. The performance evaluation tests were carried out on the test data subset, to ascertain its performance on merging instances it has never encountered before.

\begin{figure*}[!t]
    \centering
    \subfloat[Negative Rewards \label{fig:dqn_rewards_a}]{\includegraphics[width=0.5\textwidth,height=3.7cm]{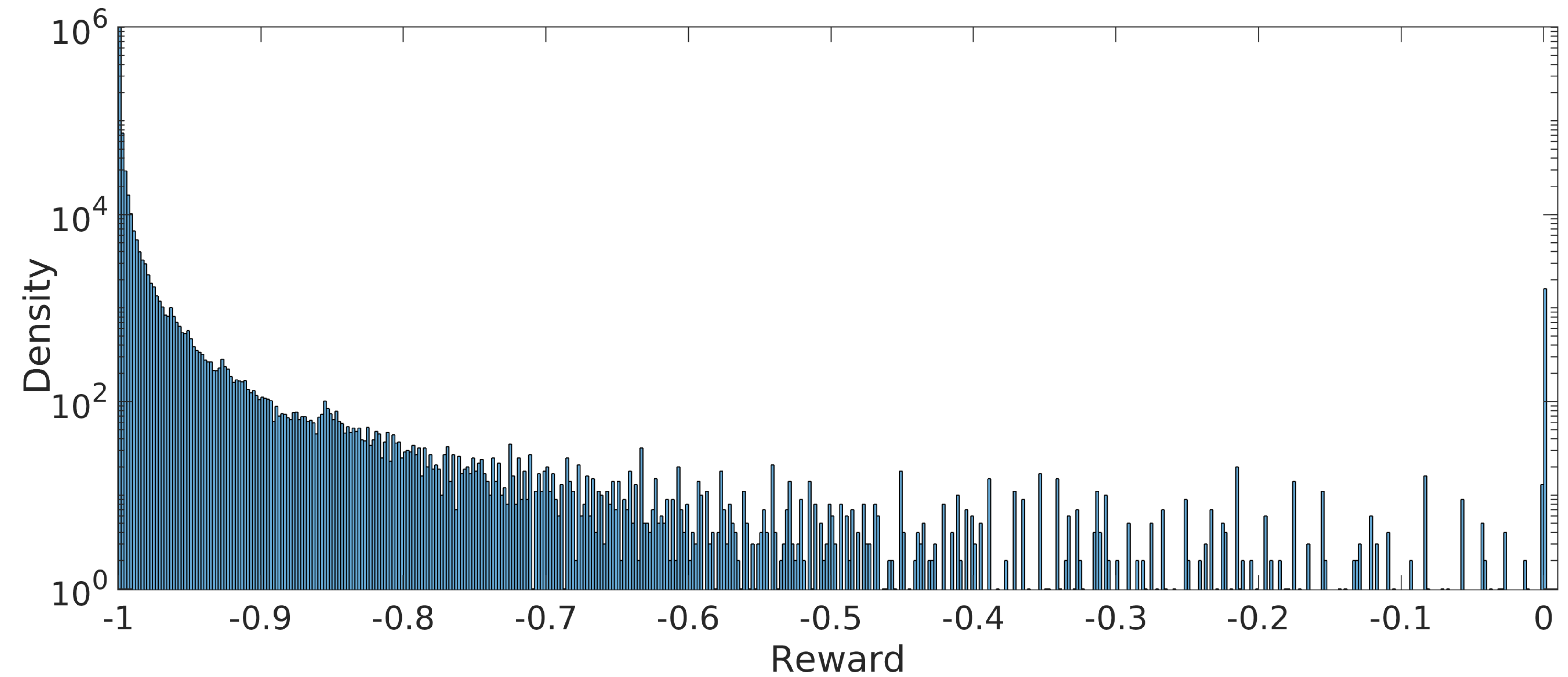}}
    \subfloat[Positive Rewards \label{fig:dqn_rewards_b}]{\includegraphics[width=0.5\textwidth,height=3.7cm]{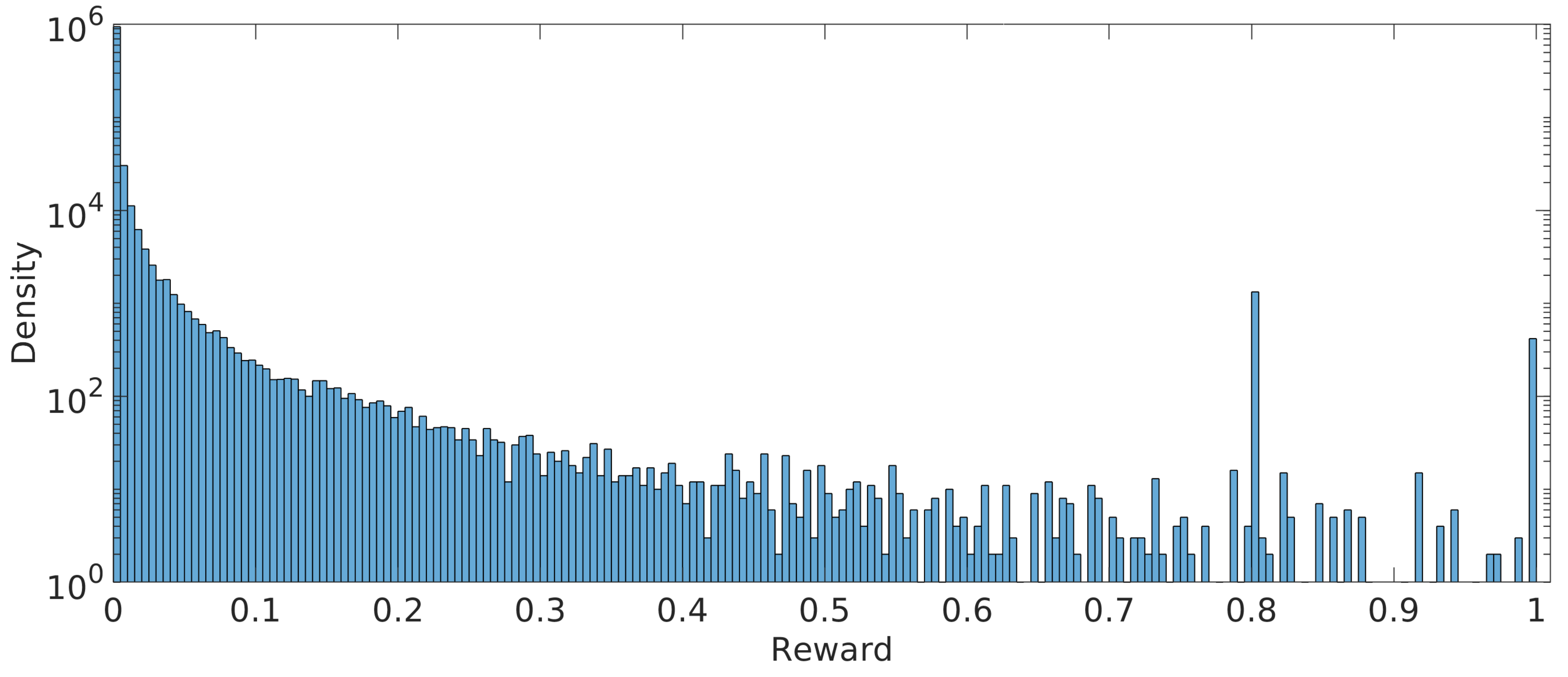}}
    \caption{Histogram for comparing assigned rewards for trajectory recommendation by \gls{dqn} agent.}
    \label{fig:comp_rew_dqn}
\end{figure*}

\begin{figure*}[!t]
    \centering
    \subfloat[Negative Rewards \label{fig:dueling_rewards_a}]{\includegraphics[width=0.5\textwidth,height=3.7cm]{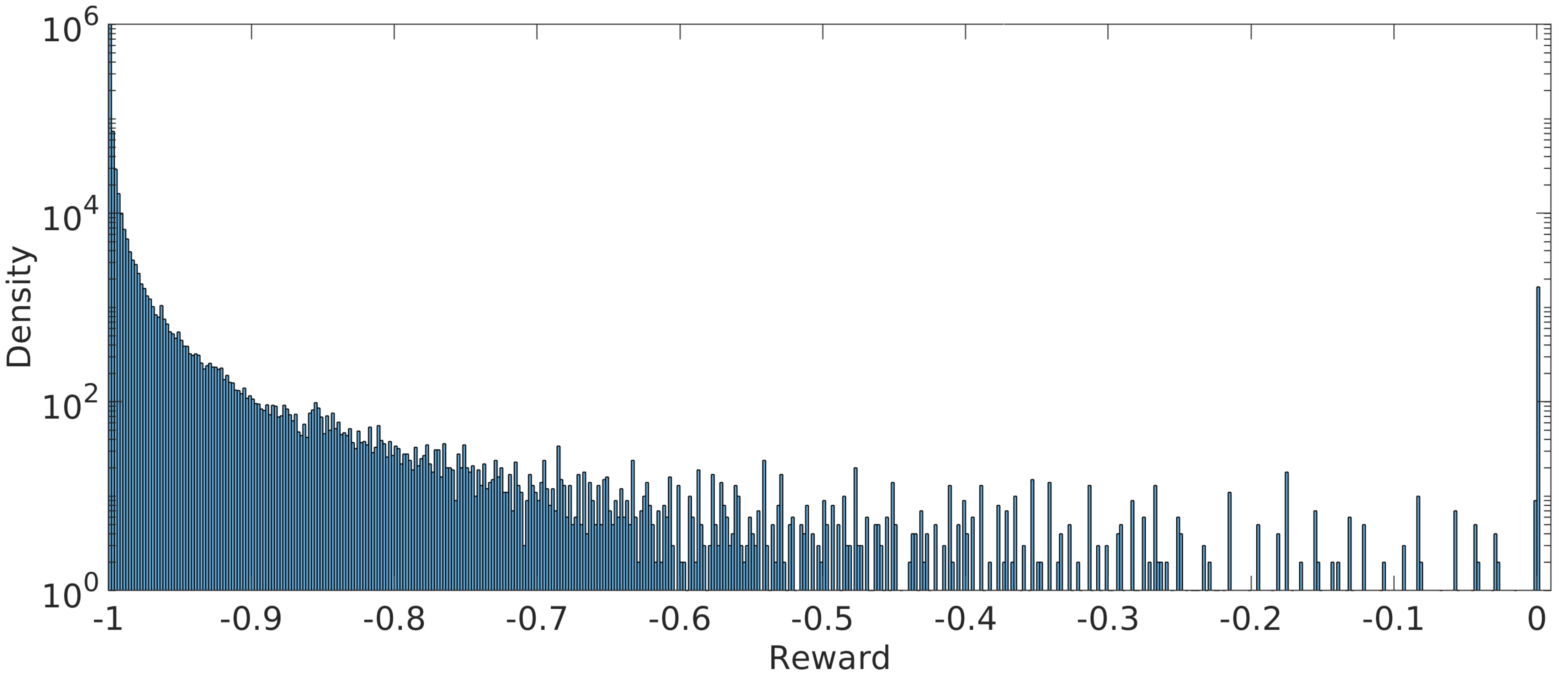}}
    \subfloat[Positive Rewards \label{fig:dueling_rewards_b} ]{\includegraphics[width=0.5\textwidth,height=3.7cm]{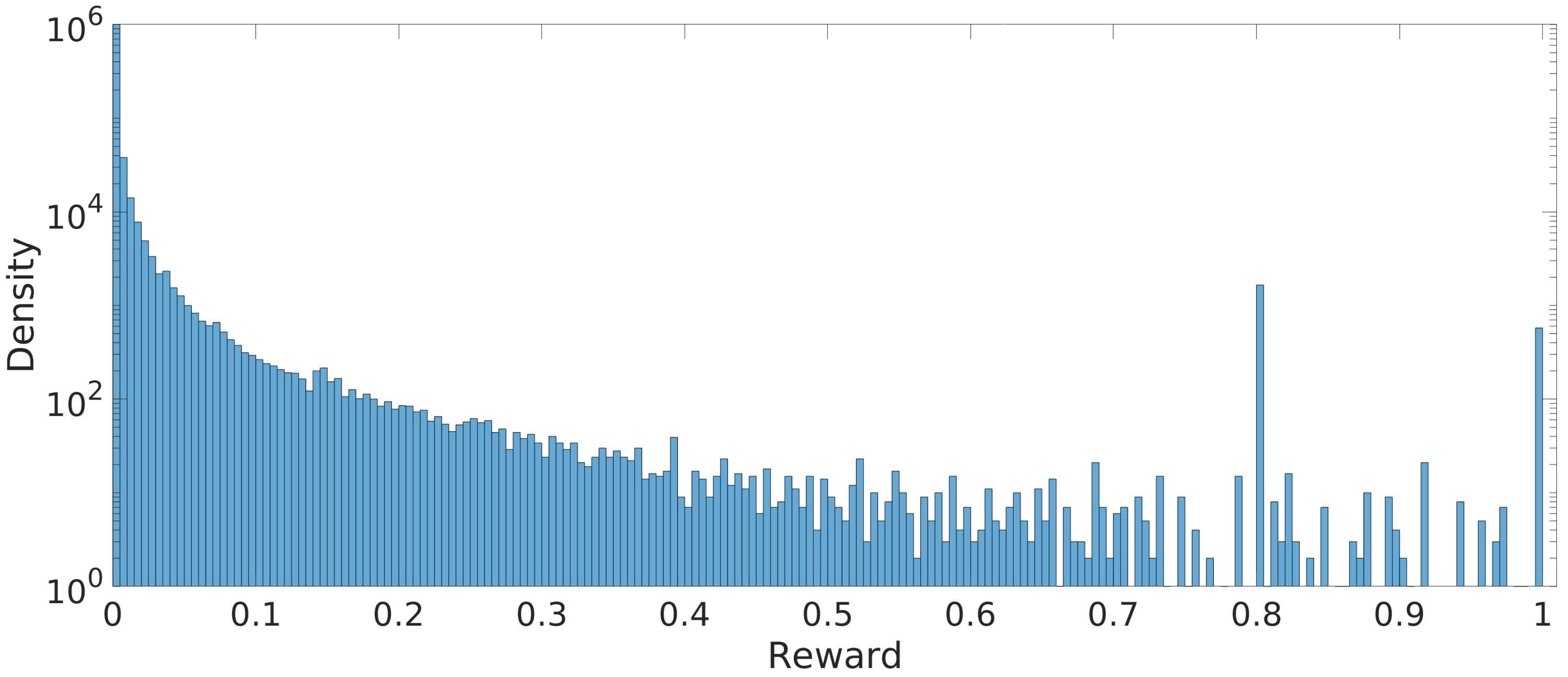}}
    \caption{Histogram for comparing assigned rewards for trajectory recommendation by \gls{dueling-dqn} agent.}
    \label{fig:comp_rew_ddqn}
\end{figure*}

Fig. $\ref{fig:comp_rew_dqn}$ and $\ref{fig:comp_rew_ddqn}$, highlight the count of positive and negative rewards assigned for each way-point in a trajectory recommendation during training time of the model. The  reward function (positive/negative) used to assign rewards directly impact the success of the trajectory recommendation. Both the models that reinforce the agent with negative rewards follow the same general shape seen in Fig. \ref{fig:dqn_rewards_a} and Fig. \ref{fig:dueling_rewards_a}, where there is an inverse proportional relation between the magnitude of the rewards and the reward the trajectory obtained, until the model reaches a successful merge obtaining a reward of $0$ where the density increases greatly, indicating a converged model. The density of rewards obtained at $0$ is  $17^{3}$ by the \gls{dueling-dqn} compared to $15^{3}$ for \gls{dqn}. On the other hand, the positive reward reinforcement of the agent also follows the same pattern, with a sudden increase in magnitude that takes place at a reward of $0.8$. Furthermore, the density of that reward surpasses the density obtained by a reward of $1$ by a minuscule factor. Notwithstanding, the \gls{dueling-dqn} still receives a higher density of greater rewards than the \gls{dqn} varying by a density of $3 \times 10^3$. Although, having the agent obtain a large density of rewards allocated at $0.8$ proves the existence of a global minima in the positive reward function that the model needed to surpass in order to obtain a successful merge, which the negative reward function did not face. Therefore, the loss and the reward assignment are used hand in hand to obtain a clearer insight of the model performance for an optimal model selection, the relatively lower loss and high successful reward density of the \gls{dueling-dqn} model utilising positive reward allocation, showcases the superiority of the model over the negative reward allocation, which was adopted for the real world lane merge tests.

\subsection{Automotive and Communication \glspl{kpi} for real vehicles}

The lane merge scenario used for the real tests consists of a test track using connected and unconnected vehicles. Four vehicles were used, three are connected: merging, following and preceding vehicle, while the fourth vehicle was unconnected. This enables the trajectory recommendation to be passed to the merging vehicle for execution, whilst giving the \textit{\gls{to}} some control over other connected vehicles in order to suggest a cooperative lane merging benefiting the entire road.  

In order to compare live \textit{\gls{to}'s} predicted trajectories and human trajectories, a preliminary test was implemented with no \textit{\gls{to}'s} interaction: several merges were performed on the test track while the KPI evaluation platform was storing the logs of those merges that occurred on the road. These stored logs are the human merges that are used to compare the predicted trajectory accuracy and the human likeness of the manoeuvre.

\subsubsection{Predicted vs Human Collected Trajectory Positioning}

\begin{figure}[!t]
    \centering
    \includegraphics[width=0.5\textwidth]{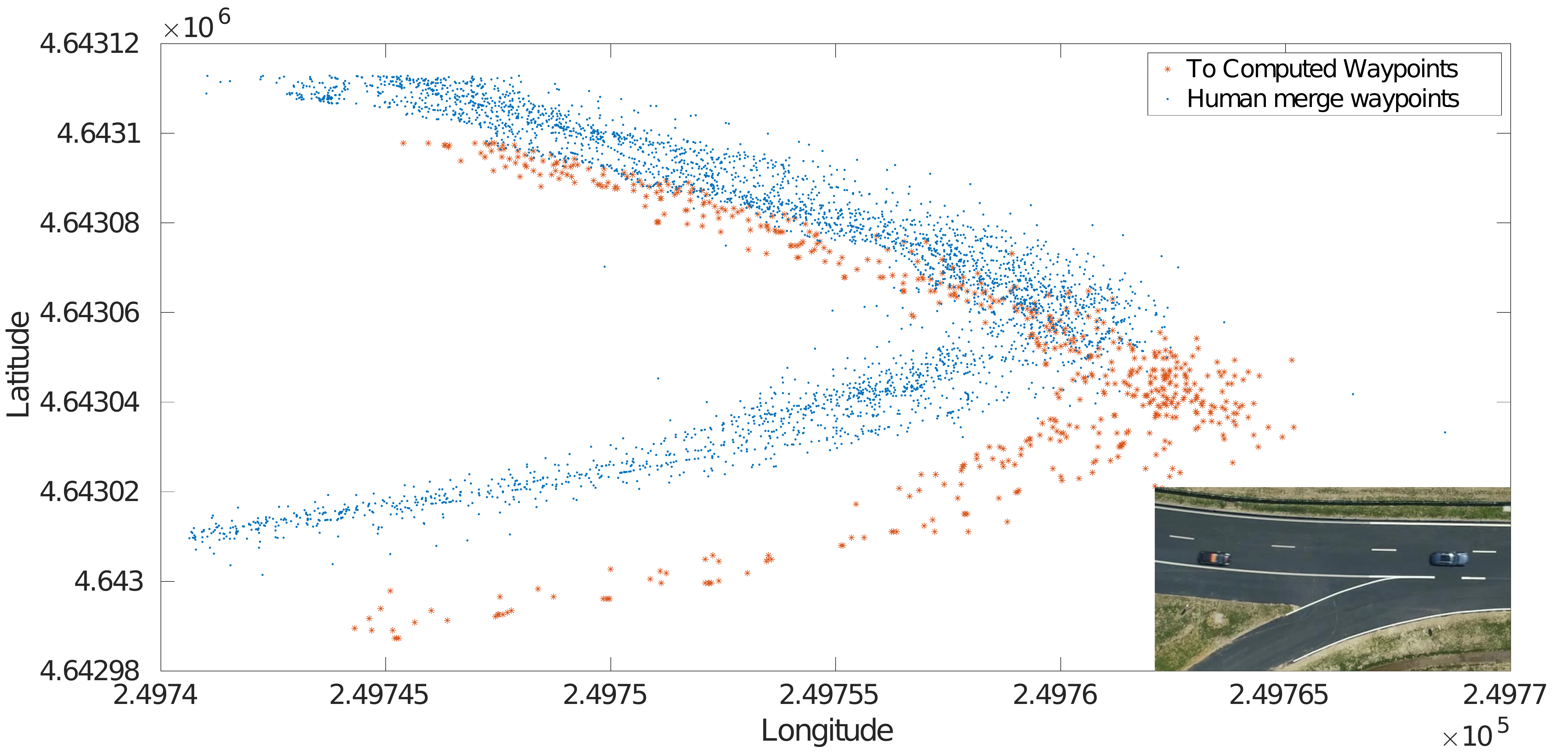}
    \caption{Human vs Computed Trajectories on Test Track}
    \label{fig:human_vs_to_real}
\end{figure}

Fig. $\ref{fig:human_vs_to_real}$ shows the latitude and longitude of the merging scenario for human and computed way points. The general shape of the merge has been detected and successfully predicted by the \textit{\gls{to}} corresponding to the road architecture, this is a good indicator that the \gls{rl} can adapt and generalise to real world scenarios that it has never encountered before. However, there is an obvious bias from the predicted way points. Since, the road information was removed from the training of the model to ensure a greater generalisation, the neural network's ability to exactly follow a trajectory that a human may undertake has been inhibited. The \textit{\gls{rl}} also required perfect synchronisation of the environment in real-time, therefore, high frequency, low latency and great precision were required to ensure that the \textit{\gls{to}} could feed the correct \gls{rud} to the \textit{\gls{rl}}. As such, the bias could stem from the minor delays the architecture incurred. The precision and accuracy of the architecture incorporating $5G$ was higher than the average \cite{5g_reliability}, this could have not been obtained by using out-of-the-box implementation of the $5G$ communication spectrum. Therefore, there is a trade off between its ability to generalise the problem to the intended behaviour that is expected to achieve, with respect to the communication architecture the model is placed in.

\subsubsection{Inter-Vehicular Distance}

\begin{figure*}[!t]
    \centering
    \subfloat[\gls{to} calculated inter-vehicle distance \label{fig:distance_real_a}]{\includegraphics[width=0.5\textwidth,height=3.7cm]{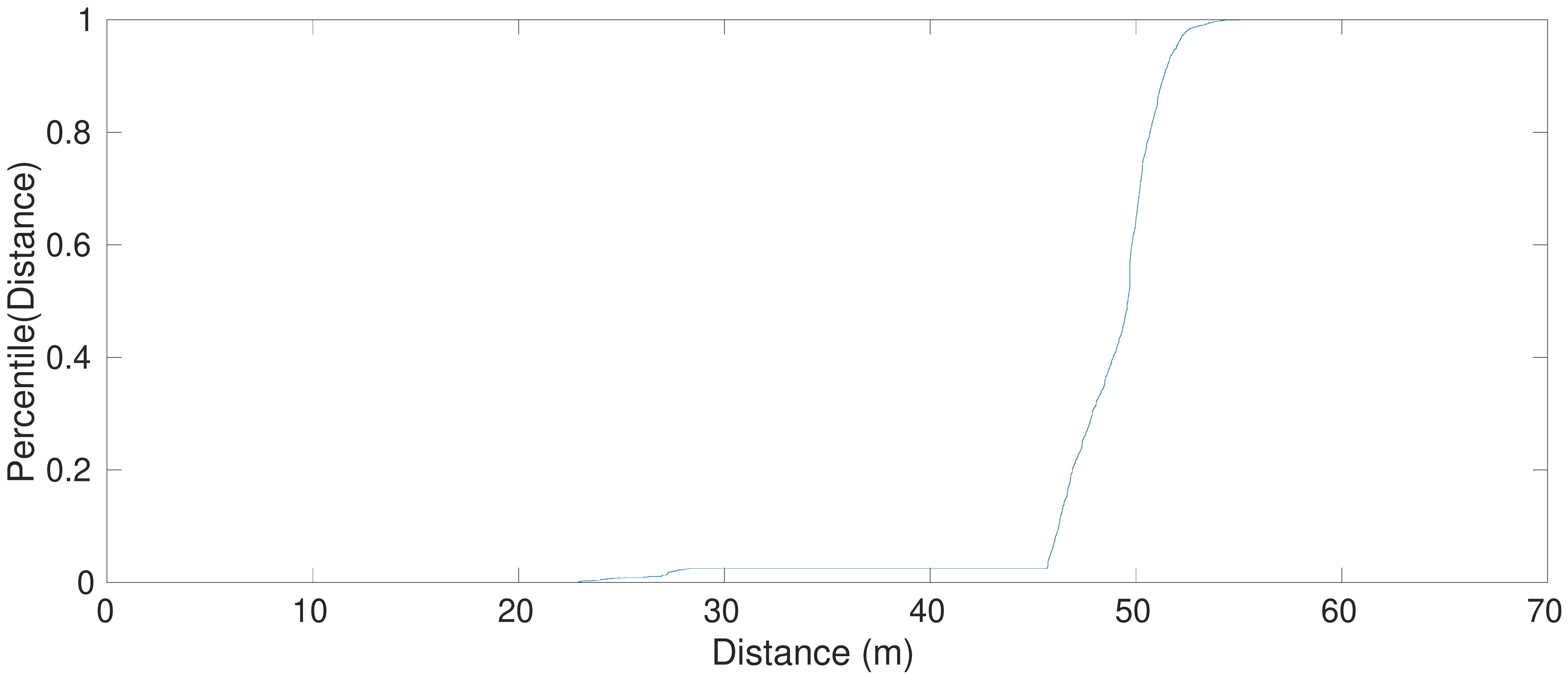}}
    \subfloat[Human calculated inter-vehicle distance \label{fig:distance_real_b}]{\includegraphics[width=0.5\textwidth,height=3.7cm]{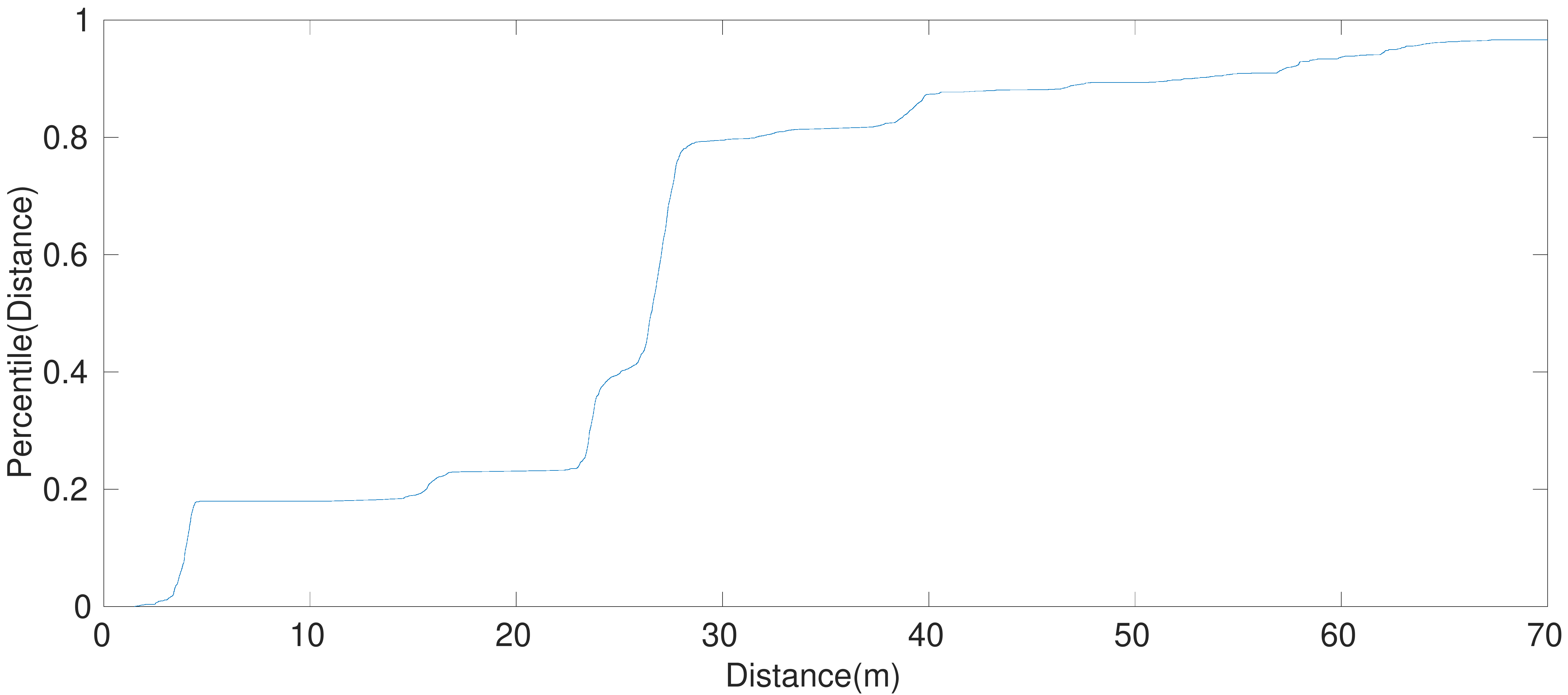}}
    \caption{\gls{ecdf} of inter-vehicle distance during merging scenario.}
    \label{fig:distance_real}
\end{figure*}

The inter-vehicle distances provide an insight on the merge of the connected vehicles in between the preceding and following vehicle. This value is mainly affected by the data fusion of the \textit{Image recognition} system and the actual connected vehicles broadcasting position. Fig. $\ref{fig:distance_real}$ presents the \gls{ecdf} of distance values recorded between vehicles laying on the same lane for the \textit{\gls{to}} and human manoeuvres. On one hand, the largest frequency of distance values lies between $48 - 60 m$ for Fig. $\ref{fig:distance_real_a}$. On the other hand, Fig. $\ref{fig:distance_real_b}$ shows that inter-vehicular distance varies greatly when the marge is undergone by humans spanning from $5 - 70 m$. This means that human merges were performed under risky situations in some cases. In counterpart, the merging car does not hinder the safety distances between the other vehicles that are presented on the road, when calculated by the \textit{\gls{to}}. In this sense, the \textit{\gls{to}} does not bias the merging distance between the preceding and following car, opting to merge approximately in between the two vehicles, to maintain the largest inter vehicle distance between the merging car and the two vehicles on the target lane. Although this does not reflect human-like driving in most cases, since the merging car can favour merging towards the preceding vehicle to allow more room for subsequent actions such as breaking, this was another design choice, in order to ensure the adaptability of the model and the approach to different lane merging scenarios, but also simplify the expected behaviour of the merging vehicle, to reduce neural network complexity and resources.

\subsubsection{Merging Acceleration}

\begin{figure}[!t]
    \centering
    \includegraphics[width=0.5\textwidth]{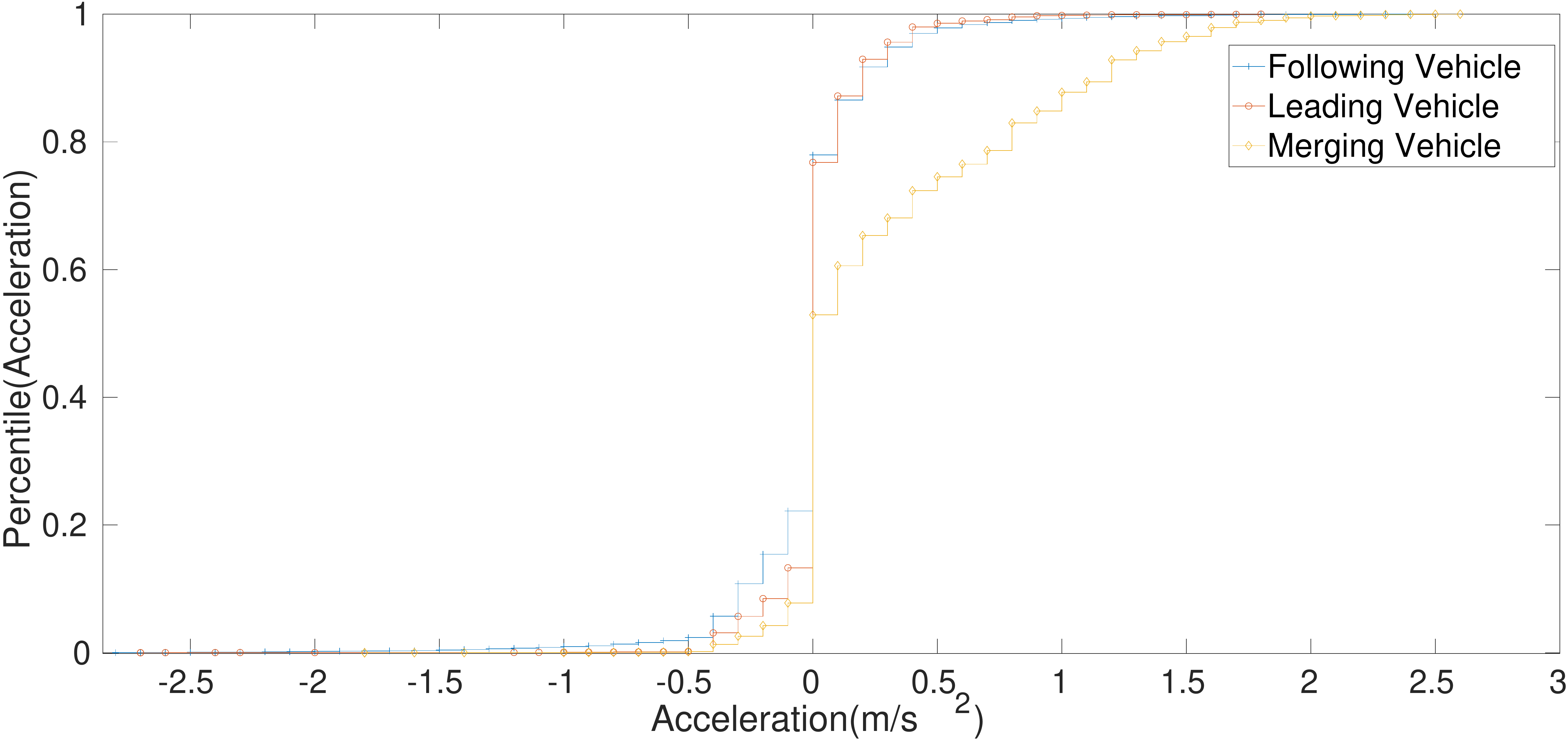}
    \caption{\gls{ecdf} of acceleration values during merging scenario.}
    \label{fig:acceleration_real}
\end{figure}

Fig.$\ref{fig:acceleration_real}$ presents the \gls{ecdf} of acceleration values against the acceleration obtained in the scenario. The acceleration values given to the merging vehicle were concentrated on speeding the vehicle up to merge in between the two vehicles from a slow lane into the target merging lane. The majority of the acceleration values lied in the range $0-2 \; m^2/s$ providing non extreme acceleration values for a merge mimicking a human driver approach to a lane merge. This means that the \textit{\gls{to}} predicted acceleration values that provides a smooth trajectory recommendation during the merge. From the following vehicle's point of view, the recommendations given had the intended purpose of slowing down the following vehicle to create a larger gap in between the vehicles on the target lane, for a safer and smoother merge experience. Consolidating the idea of a coordinated lane merge approach taken on the road. The values obtained from the \textit{V2X Gateway} displayed minor noise which further affected the speed and acceleration values recommended by the \textit{\gls{to}}. 

\subsubsection{Trajectory delivery time}

\begin{figure}[!t]
    \centering
    \includegraphics[width=0.5\textwidth]{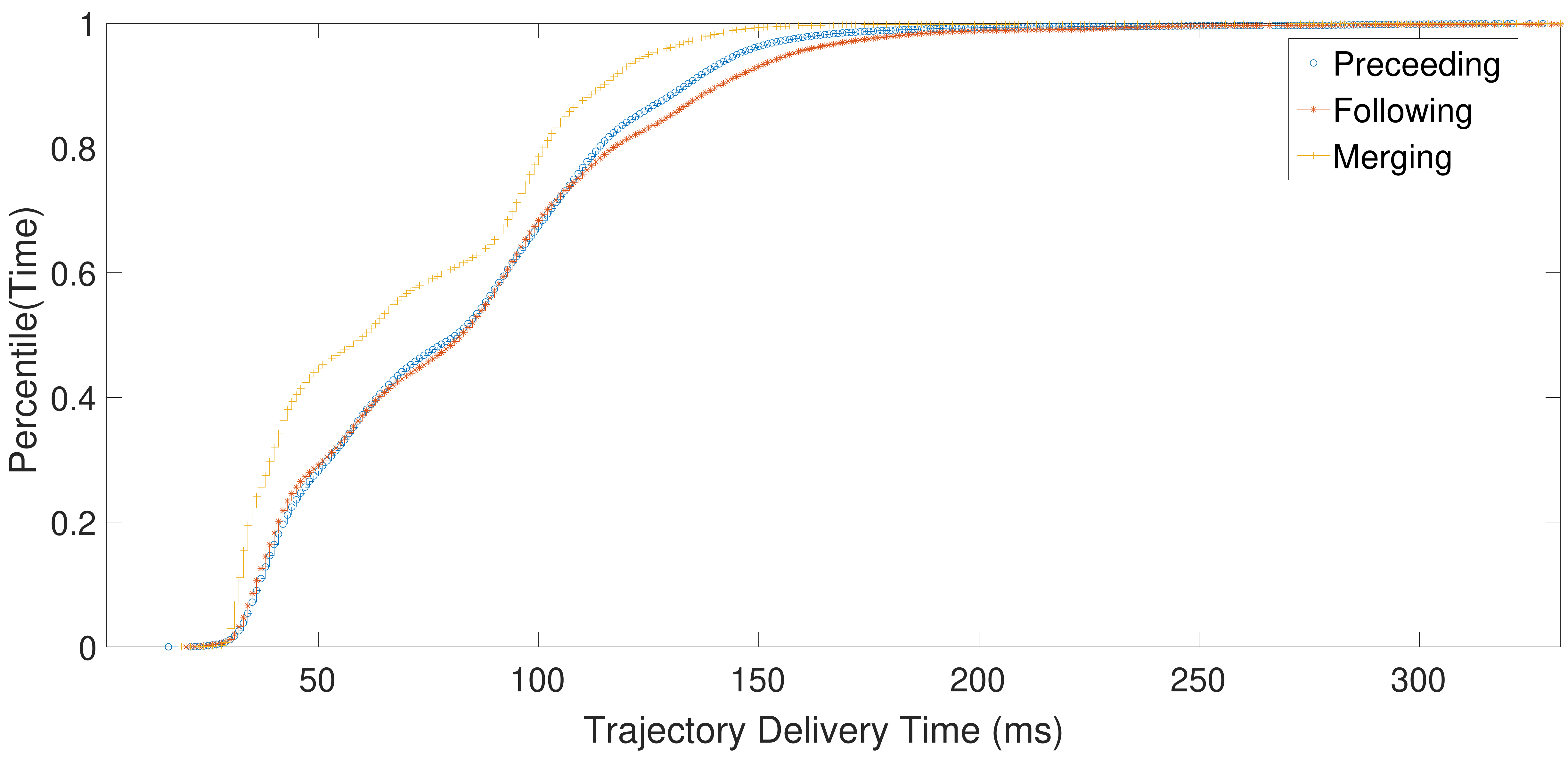}
    \caption{\gls{ecdf} of trajectory delivery time per vehicle.}
    \label{fig:all_latency}
\end{figure}

We define the \textit{trajectory delivery time} as the time it takes to deliver a manoeuvre recommendation, since the moment a \gls{rud} is first sent by the \textit{Image recognition} system (or a vehicle). The \gls{kpi} Evaluation Platform computes the time of every \gls{rud} sent from the \textit{Image recognition} system (or the vehicle itself) to the \textit{\gls{df}} and passed onto the \textit{\gls{to}} to calculate and forward a manoeuvre recommendation to the \textit{\gls{v2x}} which broadcasts it to the vehicle. Fig. $\ref{fig:all_latency}$ shows the trajectory delivery time per vehicle in which $380,000$ measurements were taken during the merging tests. It is clear that $99.9$ percentile of the measurements are under a rate of $288 ms$ (receiving the locality of the vehicle or sending the trajectory recommendation is roughly $144ms$). In terms of processing delays, the \textit{\gls{to}} manoeuvre computation is negligible in the scenario. In most of the cases, it was not possible to obtain \textit{\gls{to}'s} computation estimations due to the logging time granularity. The \textit{\gls{to}} achieved a real-time environment processing, generating safe and successful manoeuvres for vehicles in need. 

On the other hand, analysis showed that the \textit{\gls{df}} is adding a larger computational delay due to its default usage of \gls{tcp} configurations. The use of Nagle algorithm combined with delayed acknowledgements by \gls{tcp}, could result in $200ms$ of added latency according to \cite{nagle1}, which could be the source of the slight bottleneck stemming from the \textit{\gls{df}}.

%% file: conclusion.tex
In this paper, we presented a lane merge coordination model based on a centralised system. The \gls{dueling-dqn} model has been identified as the best approach compared to the \gls{dqn}, obtaining more optimal performance and providing more human-like trajectories. Predicted trajectories provided smooth driving experience of acceleration in the range of $0-2 \; m^2/s$. Future works need to be carried out in order to improve \textit{Data Fusion's} performance, processing time and transport protocol optimisation are points to be addressed.